\documentclass[letterpaper]{article} 
\usepackage[preprint]{aaai2027}  
\usepackage[hyphens]{url}  
\usepackage{graphicx} 
\usepackage{natbib}  
\usepackage{caption} 
\usepackage{algorithm}
\usepackage{algorithmic}

\usepackage{amsmath}
\usepackage{amssymb} 
\usepackage{newfloat}
\usepackage{listings}
\DeclareCaptionStyle{ruled}{labelfont=normalfont,labelsep=colon,strut=off} 
\floatstyle{ruled}
\newfloat{listing}{tb}{lst}{}
\floatname{listing}{Listing}

\usepackage{booktabs}
\usepackage{multirow}

\usepackage{amsthm}

\usepackage{pdfpages}

\theoremstyle{definition}
\newtheorem{dfn}{Definition}

\title{Diffuse to Compress: Leveraging Diffusion LMs for Lossless Compression}
\author{
    Angelo Nardone\textsuperscript{\rm 1}\corresponding,
    Paolo Ferragina\textsuperscript{\rm 1,2}
}
\affiliations{
    \textsuperscript{\rm 1} University of Pisa, Department of Computer Science, Pisa, Italy \\
    \textsuperscript{\rm 2} Scuola Superiore Sant'Anna, Pisa, Italy \\
    angelo.nardone@phd.unipi.it, paolo.ferragina@santannapisa.it
}

\begin{document}

\maketitle

\begin{abstract}
We study the problem of lossless text compression, motivated by the rapid 
growth in the collection and storage of digital textual data --- including
plain text, source code, and structured formats such as XML --- and by recent advances in neural language model-based compression. In particular, recent  LLM-based approaches, whether built on symbol-ranking pipelines or paired with a statistical compressor, have demonstrated compression ratios significantly superior to general-purpose compressors such as \texttt{zstd}, \texttt{gzip}, or \texttt{bzip} on text and code. However, these neural approaches suffer from severe throughput limitations, making them not yet practically usable.

For the first time in the context of lossless neural text compression, we introduce Diffusion Language Models (DLMs) as an alternative inference paradigm to autoregressive LLM-based approaches. We argue that replacing autoregressive LLMs with DLMs within the same compression framework could overcome the throughput bottleneck caused by their one-symbol-per-step limitation. However, achieving these improvements requires addressing algorithmic challenges introduced by applying DLMs to lossless compression, where the architecture allows the number and positions of symbols encoded at each forward pass to be decided independently. We design efficient and effective strategies to solve these challenges and evaluate them experimentally against LLM-based and general-purpose compressors on \texttt{enwik8}, a well-established textual benchmark. 

Our results show that the newly proposed DLM-based framework advances the state of the art in lossless text compression. Moreover, as DLMs are still a relatively young paradigm, recent advances toward increasingly capable and efficient models suggest substantial room for further improvements.

\end{abstract}


\section{Introduction}

Data compression is one of the most fundamental problems in Computer  Science, with direct implications for storage costs, data transfer 
speed, energy consumption, and scalability of HPC systems. The rapid growth of digital 
data --- fueled in large part by the demand for large-scale datasets to 
train and deploy AI models --- has made these concerns increasingly 
pressing. Among the data types that have seen the sharpest growth is 
textual data: plain text, source code, and structured formats such as 
XML and JSON now populate vast digital archives worldwide. Prominent 
examples include \textit{Common Crawl}~\cite{commoncrawl}, one of the 
largest open repositories of plain text web data, and \textit{Software 
Heritage}~\cite{softwareheritage:2025}, the largest archive of public source code.

A significant share of such archives falls into the category of 
\textit{cold storage}: systems optimized for long-term preservation of 
massive datasets at very low cost, where data is written once and 
retrieved rarely. In this setting, compression ratio is the dominant 
concern --- even at the cost of slower compression speed. General-purpose 
compressors, such as \texttt{zstd} or \texttt{lzma}, offer a good 
speed-ratio trade-off but fail to capture the grammatical and lexical 
regularities inherent in natural language, leaving 
substantial compression gains on the table.

This limitation, combined with the well-known connection between 
accurate language prediction and compression 
efficiency, dating back to~\cite{shannon1951prediction}, has motivated a growing body 
of work on lossless text compression based on Large Language 
Models (LLMs), such as \cite{mittu2024finezip,valmeekam2023llmzip,narashiman2024alphazip, deletang2024language}. These approaches have proven remarkably effective, 
achieving compression ratios well beyond those of general-purpose tools 
--- making them particularly promising for cold storage archives. 
However, even in this setting, where speed requirements are relaxed, their throughput remains impractically low, as inference dominates the computational cost because autoregressive LLMs are constrained to sequential symbol generation. Consequently, new neural-based architectures are needed to preserve the compression quality achieved by LLMs while providing better speed performance.

In this work, we introduce, for the first time in the context of neural 
lossless text compression, \emph{Diffusion Language Models} (DLMs) as an alternative inference paradigm to autoregressive LLM-based approaches. Unlike autoregressive LLMs, DLMs make the selection of symbols to be encoded at each inference step a design choice rather than a model constraint, allowing the number and positions of encoded symbols to be controlled independently. We therefore argue that the ability of DLMs to predict multiple symbols per forward pass may be leveraged to mitigate the architectural bottleneck imposed by one-symbol-per-step inference. However, this requires solving two main algorithmic challenges that are raised by the DLM architecture. The first one concerns the symbol commitment schedule employed by DLMs during the compression process; the second one relates to identifying effective initial contexts that may allow DLMs to avoid inefficient first-step predictions and consequently improve compression quality. We design strategies for solving both these challenges and evaluate them experimentally against LLM-based and general-purpose compressors on \texttt{enwik8}, a well-established benchmark in this setting \cite{mittu2024finezip}. Our DLM-based compressors improve compression–throughput trade-offs over prior neural approaches and open promising directions for future research.

\subsection{State of the Art}\label{sec:sota}

The idea of leveraging neural networks for lossless data compression is not 
new~\cite{mahoney2000fast, schmidhuber1996sequential}. However, the 
recent progress of neural language models has reignited interest, leading to two main 
pipelines: (i) using the neural model as a probability estimator paired with  a statistical coder, typically arithmetic 
coding~\cite{witten1987arithmetic}; or (ii) using the neural model as a 
predictor within the Symbol Ranking pipeline~\cite{shannon1951prediction}, paired with a general-purpose 
compressor.

In the former class belong approaches such as \emph{DeepZip}~\cite{goyal2018deepzip} and 
\emph{NNCP}~\cite{bellard2021nncp}, which combine RNN/LSTM-based predictors with Arithmetic Coding, thus achieving promising compression ratios at substantial computational cost. The introduction of the Transformer 
architecture~\cite{vaswani2017attention} marked a turning point, 
positioning LLM-based methods as the strongest candidates for designing probability estimators. \emph{LLMZip}~\cite{valmeekam2023llmzip} was among the 
first to combine LLMs with both Symbol 
Ranking and Arithmetic Coding. \emph{FineZip}~\cite{mittu2024finezip} 
then extended \emph{LLMZip} with parameter-efficient fine-tuning via LoRA, 
while \emph{AlphaZip}~\cite{narashiman2024alphazip} focused on smaller 
models paired with \texttt{gzip} or 
Brotli~\cite{alakuijala2018brotli}. More recently, \citet{deletang2024language} studied the use of arithmetic coding 
with three foundation models of varying scale, showing that pretrained models 
can act as compressors across text, video, and images. \citet{heurtel2024compression} 
extended this analysis to byte-level multimodal compression, showing that small 
pretrained models can outperform standard compressors, even accounting for 
model size, although they struggle with unseen modalities. Finally, \cite{nardone2026llmbased} studied 30 pretrained LLMs for source code 
compression, showing that the compression ratio--throughput trade-off depends 
strongly on model size and quantization, that small models enable high 
throughput, and that source code is even more compressible than standard 
text.

Despite their differences, all these approaches share a common limitation. 
While outperforming general-purpose compressors in compression ratio, they 
suffer from severely limited throughput. Since Transformers scale 
quadratically with context length, recent works~\cite{mittu2024finezip, 
deletang2024language} tried to address this limitation by partitioning the input 
into independently compressed chunks, enabling parallel processing across 
multiple GPUs. This strategy slightly degrades compression quality, but 
does not achieve a significant improvement in throughput, which remains in the order of kilobits per 
second~\cite{mittu2024finezip}, thus making these methods impractical even with multi-GPU parallelization.

Crucially, throughput is rarely treated as a primary metric in this setting: in fact, \citet{deletang2024language} reports no explicit timing data, and decompression 
time is never reported in any of these works. More recently, \cite{nardone2026llmbased} provided the first systematic study of the compression--throughput trade-off, showing that both metrics can be improved through proper algorithmic choices, but all of them remain constrained by the autoregressive paradigm adopted by LLMs. Closing the gap between neural and 
general-purpose compressors in throughput --- while preserving a compression 
ratio advantage --- is therefore a key challenge. We argue that achieving this goal requires an change in the compression paradigm: the throughput bottleneck is architectural, rooted in the fact that LLMs are designed to generate one symbol per forward pass --- a consequence of the autoregressive next-symbol prediction paradigm that cannot be overcome by optimizing existing pipelines.

\subsection{Our Contribution}
Motivated by these limitations, we advance neural lossless compression with the following contributions.
\begin{enumerate}
    \item We introduce, for the first time in the literature, the use of Diffusion Language Models (DLMs) for lossless text compression, establishing a new {\em non}-autoregressive compression paradigm. Unlike autoregressive LLMs, DLMs make the selection of symbols to be encoded at each inference step a design choice rather than a model constraint, allowing the number and positions of encoded symbols to be controlled independently. We instantiate this framework using \emph{LLaDA}~\cite{nie2025large}, one of the first masked DLMs released on Hugging Face, which is competitive with 
    autoregressive LLMs such as LLaMA~3~8B \cite{grattafiori2024llama} in language modeling.
        
    \item Our paradigm change needs to solve two novel algorithmic challenges raised by the DLM architecture. The first concerns the symbol commitment schedule employed during the compression process; the second hinges on the ability to provide arbitrary initial contexts to avoid inefficient first-step predictions, reduce the number of symbol-generation stages, and thus improve compression performance. We design and evaluate various strategies for solving both challenges efficiently (cfr. throughput) and efficaciously (cfr. compression ratio), thus improving prediction quality while avoiding wasteful inference time.
    
    \item We experimentally evaluate our DLM-based compression framework against prior LLM-based approaches and general-purpose compressors on \texttt{enwik8}, a standard benchmark shared with prior work~\cite{mittu2024finezip}. Our results corroborate the theoretical analysis and confirm that our DLM-based compressors advance the state of the art in neural lossless compression, coming closer to usable end-to-end neural compressors. Moreover, DLMs are still a relatively young paradigm, and recent advances already suggest substantial room for further improvements.

\end{enumerate}

\section{Theoretical Background}

\subsection{General-Purpose Compressors}\label{sec:gpcf}

Modern general-purpose compressors follow three main algorithmic paradigms, differing in how they identify and encode redundancy. Statistical compressors operate through a modeling phase, where symbol probabilities are estimated or fixed in advance, followed by a coding phase using entropy coders such as Huffman coding~\cite{huffman1952method}, Arithmetic Coding~\cite{rissanen1979arithmetic}, or Asymmetric Numeral Systems~\cite{duda2009ans}. We adopt \textit{range Asymmetric Numeral Systems} (rANS), which combines near-optimal entropy coding efficiency with high throughput. Dictionary-based compressors exploit repeated substrings through dictionary references, typically based on \textit{LZ77}~\cite{ziv1977universal}. We consider three representative implementations: \texttt{gzip}~\cite{deutsch1996deflate}, \texttt{lzma}~\cite{pavlov2024lzma}, and \texttt{zstd}~\cite{collet2018zstandard}. Finally, block-sorting compressors apply the reversible \textit{Burrows-Wheeler Transform} (BWT)~\cite{burrows1994block}, followed by \textit{Move-to-Front}~\cite{bentley1986locally}, \textit{Run-Length Encoding}~\cite{ferragina2023pearls}, and an entropy coder. We use the representative implementation \texttt{bzip2}~\cite{seward1996bzip2}.

\subsection{Neural Language Model-Based Compressors}

Neural language model-based compressors specialize on textual data, 
achieving compression ratios well beyond general-purpose tools. They 
integrate into two main pipelines: feeding predicted distributions to a 
statistical coder, or using them within the \textit{symbol ranking} 
pipeline.

\paragraph{Predictive Approaches.} These are statistical approaches mainly solving the \textit{next-symbol prediction} task, which is defined as follows.

\begin{dfn}\label{dfn:nsp}
Let \mbox{$S = s_1, \dots, s_n$} be a sequence of symbols over an alphabet $\Sigma$, and let $W_{t,k} = s_{t-k}, \dots, s_{t-1}$ denote the window (context) of $k$ 
symbols preceding position $t$ in S. The \textit{next-symbol prediction} 
estimates, for each $t$, a distribution $p(\cdot \mid W_{t,k})$ over 
$\Sigma$ so as to maximize $p(s_t \mid W_{t,k})$.
\end{dfn}

The classical instantiation of this approach is \textit{PPM}~\cite{cleary1984ppm, 
moffat1990ppm, cleary1997unbounded,ferragina2023pearls}, which uses explicit 
variable-length context matching over the prefix $W_{t,t-1}$. Large Language Models (LLMs) represent a 
far more powerful instantiation: trained to optimize the 
\textit{language modeling} objective --- by assigning a joint probability 
distribution to sequences of symbols via the chain rule:
\[
    p_\theta(S) = \prod_{t=1}^{n} p_\theta(s_t \mid s_1, \dots, s_{t-1})
\]
--- they reduce exactly to next-symbol prediction at every position. 
Unlike PPM, LLMs learn contextual representations and capture long-range statistical dependencies beyond what exact context matching can 
achieve~\cite{vaswani2017attention}. This makes them ideal for implementing the 
modeling phase of a predictive compressor: their predicted distributions 
over the next symbol are fed directly to a statistical coder 
such as Arithmetic Coding. This is exactly what a robust line of research has investigated recently, starting with the papers of~\cite{deletang2024language, valmeekam2023llmzip, 
mittu2024finezip}, and yielding compression ratios well beyond general-purpose 
compressors. However, this comes at the cost of extremely low throughput.

\paragraph{Symbol Ranking.}
Rather than feeding distributions to a statistical coder, symbol ranking
orders symbols by likelihood, as originally suggested by~\cite{shannon1951prediction}.

\begin{dfn}[Symbol Ranking]\label{def:sr}
Let $S = s_1, \dots, s_n$ be a sequence of symbols over an alphabet $\Sigma$, and let $p(\cdot \mid W_{t,k})$ be the next-symbol 
prediction distribution (Definition~\ref{dfn:nsp}) at position $t$. This induces an 
ordered list of symbols $\sigma_0, \dots, \sigma_{|\Sigma|-1}$ such that 
$p(\sigma_0 \mid W_{t,k}) \geq \dots \geq p(\sigma_{|\Sigma|-1} \mid 
W_{t,k})$. This way, the next symbol $s_t$ is assigned the rank
\[
    \text{Rank}(s_t \mid W_{t,k}) = j \iff s_t = \sigma_j.
\]
Applying this at every position defines the transformation from the sequence of symbols $S$ to the sequence of ranks (integers) $R = r_1, \dots, r_n$, with $r_t = \text{Rank}(s_t \mid W_t)$.
\end{dfn}

Since symbol ranking is a transformation rather than a compressor, the rank sequence $R$ must be compressed
by a downstream general-purpose compressor. \citet{fenwick1997symbol}
observed that this approach is related to block-sorting
compressors, as both produce skewed output distributions.

Prior LLM-based works~\cite{valmeekam2023llmzip, mittu2024finezip, 
narashiman2024alphazip} have employed general-purpose compressors such 
as \texttt{gzip}~\cite{deutsch1996deflate}, 
\texttt{zlib}~\cite{deutsch1996rfc1950}, 
\texttt{bzip2}~\cite{seward1996bzip2}, and adaptive Huffman 
coding~\cite{gallager1978variations}, showing that symbol ranking yields 
slightly worse compression ratio than Arithmetic Coding, but higher 
throughput, as compressing integer ranks with general-purpose compressors 
is considerably faster than the application of Arithmetic Coding over model-predicted distributions.

\section{Diffusion-Based Compression}


Rather than further optimizing existing LLM-based pipelines, we address their limitations by introducing Diffusion Language Models (DLMs) as an alternative backbone for neural compression. This substitution preserves the compression pipeline while introducing new algorithmic challenges, whose solution enables a new compression paradigm with state-of-the-art compression--throughput trade-offs.

\subsection{Diffusion Language Models}

Diffusion Language Models (DLMs) are a class of non-autoregressive 
Transformer-based sequence models that generate text through an iterative denoising 
process, in contrast to autoregressive generation. 
Rather than parameterizing $p_\theta(S)$ as an ordered product of 
conditional probabilities, DLMs model a forward corruption process and 
learn to invert it. Interest in DLMs is growing rapidly, with recent 
models such as \textit{DiffusionGemma}~\cite{diffusiongemma2026} 
demonstrating throughput far exceeding autoregressive alternatives.

Different forward corruption processes give rise to families of DLMs 
with different inference properties~\cite{vonrutte2025scaling}. We focus 
on \textit{masked diffusion}, in which subsets of symbols are replaced 
by \texttt{[MASK]} and the model reconstructs them from the remaining 
context.

\begin{dfn}[Masked Diffusion Language Model]\label{dfn:dlm}
A \textit{masked diffusion language model} is a parametric function 
$p_\theta(\cdot \mid \cdot)$ that, given a partially observed sequence 
$\tilde{S} = \tilde{s}_1, \dots, \tilde{s}_n$ over $\Sigma \cup 
\{\texttt{[MASK]}\}$, estimates in a single forward pass a probability 
distribution over $\Sigma$ for each masked position
$t \in \mathcal{M}= \{x : \tilde{s}_x = \texttt{[MASK]}\}$:
\[
    p_\theta(\cdot \mid \tilde{S}) \in \Delta(\Sigma)^{|\mathcal{M}|},
\]
where $\Delta(\Sigma)$ denotes the probability simplex over $\Sigma$.
\end{dfn}

Predictions are \emph{not conditioned on a 
fixed left-to-right prefix}: each masked position attends to all available 
unmasked positions, regardless of order. Masked diffusion thus produces, 
at each inference step, a well-defined probability distribution over all 
masked positions, enabling masked DLMs to replace LLMs in compression 
pipelines while avoiding their sequential decoding bottleneck. While other diffusion formulations, such as \textit{uniform diffusion}, exist, they do not naturally provide the conditional token distributions required by entropy coding and therefore require non-trivial adaptations.

\begin{figure*}[t]
\centering
\includegraphics[width=2\columnwidth]{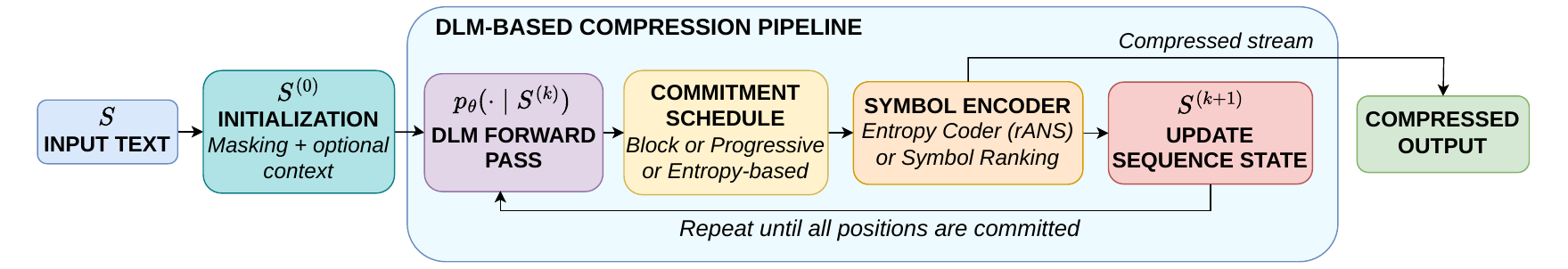}
\caption{Iterative DLM-based pipeline for lossless text compression.}
\label{fig:pipeline}
\end{figure*}

\subsection{Diffusion-Based Compression Pipeline}
\label{sec:ar_vs_nar}

Given a sequence of symbols $S = s_1, \dots, s_n$ over $\Sigma$ to be compressed, 
a diffusion-based pipeline starts from the fully masked state $S^{(0)} = \texttt{[MASK]}, \dots, \texttt{[MASK]}$, and iteratively reconstructs the sequence over $K$ inference steps:
\[
    S^{(k+1)} = f_\theta\!\left(S^{(k)}\right), \qquad k = 0, \dots, K-1.
\]
At each step, the DLM produces conditional probability distributions over 
all currently masked positions simultaneously, with no ordering constraint 
imposed by the model itself. The commitment schedule selects a subset of positions, and the corresponding original symbols are encoded using their predicted distributions.
As in LLM-based compression pipelines, these distributions can be directly 
provided to a statistical coder (e.g., rANS) or 
converted into symbol rankings. Once encoded, the original symbols are 
revealed in the sequence state and become additional context for subsequent 
inference steps, until the full sequence is reconstructed.

The number of inference steps $K$ is freely tunable and depends on the 
schedule used to select symbols at each step. For small $K$, many positions 
are committed per forward pass, reducing inference cost but providing less 
context for future predictions. Conversely, larger $K$ values reveal 
symbols gradually, improving prediction accuracy at the cost of 
additional inference time. Therefore, the commitment schedule,
including \textit{which} symbols to commit and \textit{how many} at each step, 
governs the compression ratio--throughput trade-off.

Prior work on LLM-based compressors has shown that their computational cost is dominated by model inference and can be characterized by the 
number of sequential forward passes for a fixed context length~\cite{mittu2024finezip, 
valmeekam2023llmzip}. While the absolute cost of each pass depends on factors such as model size, quantization, and hardware, when comparing DLMs and LLMs of similar size, the number of inference passes remains the dominant factor determining throughput.


In the autoregressive setting, the factorization of $p_\theta(S)$
requires computing $p_\theta(\cdot \mid s_1,\dots,s_{t-1})$ for each
$t=1,\dots,n$ sequentially, resulting in $n$ forward passes. Diffusion-based compression instead requires only $K$ evaluations, where $K$ is tunable through the commitment schedule, transforming sequential depth from a model constraint into a design choice that, actually, opens a spectrum of compression ratio–throughput operating points, and consequent trade-offs, unavailable to autoregressive models.

\subsection{Symbol Commitment Schedules}
\label{sec:inference}

Unlike LLMs, DLMs decouple inference from symbol encoding order, allowing the commitment schedule to be optimized independently of the model. This introduces a new algorithmic challenge: the choice of which symbols to encode at each inference step and how many to commit. We 
denote by $u_k$ the number of symbols committed at inference step $k$, 
with coverage constraint $\sum_{k=1}^{K} u_k = n$. The two extreme cases 
--- one-shot ($K=1$) and autoregressive ($K=n$) --- represent undesirable 
operating points: the former requires predicting the full string $S$ in one-shot from an almost empty 
context, while the latter recovers the sequential decoding bottleneck. We study and design three commitment strategies of increasing sophistication between these two extremes.

\paragraph{1. Uniform Block Schedule.} Given a fixed hyperparameter $K$, we commit $u = \lfloor n/K \rfloor$ 
symbols per inference step in contiguous left-to-right blocks. This 
schedule generalizes the autoregressive 
setting by committing $u$ consecutive symbols at each step.

\paragraph{2. Progressive Schedule.} This schedule is controlled by the initial budget $u_1$ and growth ratio
$r$. Early inference steps operate with limited revealed context and thus may 
yield less reliable predictions; later steps have access to additional 
context and can therefore commit more symbols. We deploy this observation as follows: as in uniform block 
scheduling, symbols are committed left-to-right, but according to a 
\emph{geometric schedule}: $u_k = \left\lfloor u_1 \cdot r^{\,k-1} \right\rceil$.
When $r=1$, this reduces to uniform block scheduling.

\paragraph{3. Entropy-Based Schedule.}
This schedule is controlled by an entropy threshold $\tau$ and the minimum 
number of committed symbols $m$. Unlike the two strategies above, it 
exploits the bidirectional context of DLMs, that is, rather than committing symbols 
left-to-right, it lets the model decide \emph{which} positions to reveal 
based on prediction confidence. More precisely, for each currently masked position $j$, we 
define $p_j^{(k)}(v)=p_\theta(v \mid S^{(k-1)})$ and compute the entropy
\[
    H_j^{(k)}=-\sum_{v\in\Sigma}p_j^{(k)}\!(v)\: \log{p_j^{(k)}\!(v)}.
\]
At each step, we commit all positions with entropy $H_j^{(k)} \leq \tau$, namely, positions whose predictions are "almost surely" correct according to the parameter $\tau$. If 
fewer than $m$ positions satisfy this criterion, we then commit the 
$m$ positions with the lowest entropy, regardless of the threshold. 

\smallskip
All strategies are deterministic given their hyperparameters and therefore 
require no additional side information beyond the compressed file header, 
allowing the decompressor to reproduce the same commitment decisions.

\subsection{Initial Context Strategies}
\label{sec:initial}

Existing LLM-based compression pipelines start encoding text from an empty context \cite{narashiman2024alphazip,mittu2024finezip,valmeekam2023llmzip,deletang2024language}. However, this leads to poor predictions in the first inference step: without sufficient context, the model produces uncertain distributions that waste computation and hurt compression quality. This limitation is particularly relevant when multiple symbols are committed at the first step. Providing initial context could mitigate this issue, but autoregressive models are restricted to left-to-right prefixes. Conversely, DLMs allow any subset of symbols to be exposed before the first inference step, creating an optimization dimension unavailable to LLM-based compressors.

We study initial context by leaving $c$ symbols unmasked and storing them in the compressed file. While $c=0$ recovers the fully masked approach and large $c$ approaches direct storage, intermediate values trade conditioning information for storage overhead. Since DLMs attend bidirectionally, \emph{any} subset of positions can serve as context. We instantiate this optimization problem through four context selection policies, from simple baselines to model-dependent strategies.

\paragraph{1. Contiguous Context.}
The first $c$ symbols are left visible, and the rest are masked. This is 
the classical prefix-based initialization, also applicable to LLM-based 
pipelines.

\paragraph{2. Random Context.}
$c$ arbitrary positions are selected uniformly at random and left unmasked. This 
provides a simple non-contiguous baseline that ignores structural 
information.

\paragraph{3. Frequency-Based Context.}
Token frequency, derived from the tokenizer vocabulary statistics, reflects how 
often symbols appear in the training distribution: frequent symbols are less 
informative, as they are easier to predict. We exploit the frequency statistics 
of the adopted tokenizer by leaving the $c$ least frequent symbols in $S$ 
unmasked, as rare tokens provide richer conditioning information.

\paragraph{4. Attention-Based Context.}
Two complementary forward passes --- one with even-indexed positions 
masked and one with odd-indexed positions masked --- estimate the 
importance of each position through the attention it receives from masked 
tokens, aggregated via \textit{Attention Rollout}~\cite{abnar2020quantifying}. 
The $c$ highest-scoring positions are selected as initial context. This strategy requires two additional inferences during compression, 
increasing encoding time, but the selected context is stored in the file 
header and reused during decompression without additional cost.

\section{Experimental Methodology}

\paragraph{Evaluation Setup.}
We evaluate our approach on \texttt{enwik8}, a standard benchmark for 
lossless text compression, also adopted by previous LLM-based compression 
works such as \cite{mittu2024finezip}. Following prior work, we perform extensive ablation experiments on the first 1 MB of \texttt{enwik8}
and evaluate final comparisons on its first 10 MB to ensure consistency with 
previous approaches.

We compare against neural compression methods based on LLMs, including 
\cite{valmeekam2023llmzip,mittu2024finezip}, as well as 
general-purpose compressors including \texttt{gzip}~\cite{deutsch1996deflate},
\texttt{lzma}~\cite{pavlov2024lzma}, 
\texttt{zstd}~\cite{collet2018zstandard}, and 
\texttt{bzip2}~\cite{seward1996bzip2}. For a controlled comparison with 
autoregressive models, we additionally evaluate \emph{Llama-3.1-8B} 
\cite{grattafiori2024llama} using our pipeline, without providing any initial 
context and reconstructing each sequence autoregressively.

We report compression ratio, measured as compressed size over original size, 
and (de)compression throughput in MB/s. All experiments are conducted on an NVIDIA DGX H100 
server equipped with H100 GPUs.

\paragraph{Implementation Details.}
Our experiments use \emph{LLaDA}~\cite{nie2025large}, a masked diffusion
language model, as the DLM backbone. The choice of LLaDA is not intrinsic to
our framework, but is motivated by its being one of the few publicly available
masked DLMs on Hugging Face and by its competitive language modeling
performance. The model and tokenizer are accessed through Hugging Face
Transformers. The input text is tokenized using the LLaDA tokenizer and divided
into non-overlapping windows of fixed size $w$. This parameter is configurable
and stored in the compressed file header to enable deterministic decoding.
Longer windows provide richer context and can improve compression quality, but
increase the computational cost due to the quadratic scaling of Transformers
with sequence length. Non-overlapping windows also enable parallel processing
across multiple GPUs. The resulting windows are processed by our iterative pipeline shown in Figure~\ref{fig:pipeline}.

The tokens are encoded using either (i) rANS~\cite{duda2009ans} 
for the statistical coding approach, or (ii) Symbol Ranking, where integer 
ranks are encoded using byte-aligned variable-length integer (\texttt{varint}) encoding  and subsequently compressed using the general-purpose compressors described above. A list of hyperparameter configurations explored is provided in the supplementary material (Table~2).

\section{Results}


Our experiments show that the proposed approach not only introduces a new compression--throughput trade-off, but also substantially improves throughput over prior neural compressors while preserving competitive compression ratios, making DLM-based compression a promising step toward practical end-to-end usability.

\begin{figure}[t]
    \centering
    \includegraphics[width=\linewidth]{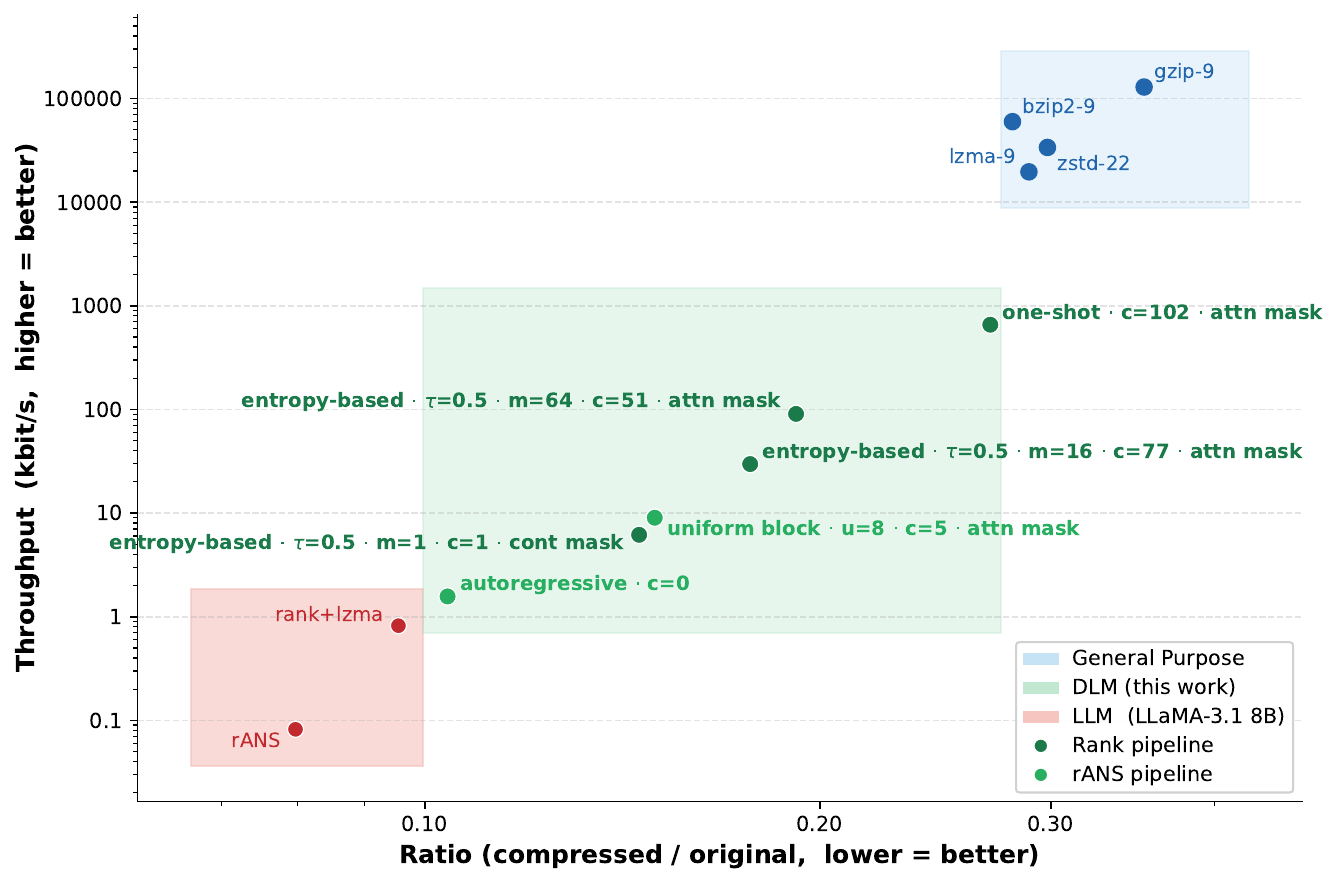}
    \caption{Compression ratio versus compression throughput (log scale) on the first 1\,MB of \texttt{enwik8}. Blue, red, and green regions denote the operating ranges of general-purpose compressors (blue), LLM-based pipeline (red), and our novel DLM-based approach (green), respectively.}
    \label{fig:scatter_overview}
\end{figure}

Figure~\ref{fig:scatter_overview} (where $y$-axis is on a log scale) summarizes the results of our DLM-based compression pipeline (green region), compared with general-purpose compressors (blue region) and an LLM-based implementation of our pipeline (red region). The three approaches occupy clearly distinct regions of the throughput--compression ratio space. General-purpose compressors achieve the lowest compression ratios but provide up to six orders of magnitude higher throughput than the LLM-based pipeline. Conversely, the LLM-based approach achieves the best compression ratios, with the statistical rANS pipeline providing better compression at the cost of lower throughput than Symbol Ranking. This confirms the current performance gap already observed in the literature \cite{mittu2024finezip,deletang2024language,nardone2026llmbased}. So, here comes the novelty and contribution of our DLM-based proposal that enables a substantially broader trade-off depending on the adopted "symbol committed schedule" and "initial context strategy", described in the previous sections. In the autoregressive configuration, which actually mimics LLMs via DLM architectures, our DLM-based pipeline achieves compression and throughput comparable to the LLM-based Symbol Ranking pipeline, showing that \emph{LLaDA}-8B remains competitive with \emph{Llama-3.1-8B} when used autoregressively. However, increasing the number of tokens generated per inference step progressively shifts the operating point of DLMs toward higher throughput, at the cost of a slightly worse compression ratio. In particular, our DLM-based approach achieves up to four orders of magnitude higher throughput than LLM-based approaches while still outperforming general-purpose compressors in compression ratio. Actually, the DLM-based compression pipeline can be broadly configured to favor either compression ratio or throughput depending on the underlying application requirements. These figures become even more interesting in the light of the current developments and potentially appealing progress underlying the design of more and more efficient DLMs, as shown in \cite{anagnostidis2025flexidit,kim2025autoregressive,diffusiongemma2026,vonrutte2025scaling}.

\begin{table}[!b]
\centering
\begin{tabular}{lccc}
\toprule
\textbf{Pipeline} & \textbf{CR} & 
\textbf{C-Thr.} & \textbf{D-Thr.} \\
& & \textbf{(kbit/s)} & \textbf{(kbit/s)} \\
\midrule
\multicolumn{4}{l}{\textit{Prior Work (teacher forcing, from \cite{mittu2024finezip})}} \\
LLMZip (AC)    & \textbf{0.079} & 0.09         & ${\ll}$\,0.09 \\
LLMZip (symb)  & 0.116          & 0.09         & ${\ll}$\,0.09 \\
 FineZip (AC)   & 0.079          & 0.10         & ${\ll}$\,0.10 \\
 FineZip (symb) & 0.128          & \textbf{5.33} & ${\ll}$\,5.33  \\
\midrule
\multicolumn{4}{l}{\textit{Our DLM-based Pipeline}} \\
 DLM (rANS) & \textbf{0.127}  & 1.01 & 1.01 \\
 DLM (symb) & 0.176 & 11.23 & 11.23 \\
 DLM (symb) & 0.223 & 44.34 & 44.35 \\
 DLM (symb) & 0.246 & \textbf{165.79} & \textbf{166.12}  \\
\bottomrule
\end{tabular}
\caption{Results on the first 10\,MB of \texttt{enwik8}. CR = compression ratio; C/D-Thr.\ = compression/decompression throughput (kbit/s). Prior LLM results are copied from~\citet{mittu2024finezip}; decompression throughput is not reported in that paper but, as commented in text, it is surely ${\ll}$\,C-Thr.}
\label{tab:main_results}
\end{table}

Let us now dig more into the technical details of our proposal and compare it with prior work. Table~\ref{tab:main_results} reports the results of LLMZip~\cite{valmeekam2023llmzip} and FineZip~\cite{mittu2024finezip} on the first 10MB of \texttt{enwik8}, as reported by \citet{mittu2024finezip}, for their two final token compression methods: arithmetic coding (AC) and Symbol Ranking (sym). Our approach provides  up to four orders of magnitude throughput improvements over prior LLM-based compressors, while maintaining a close compression ratio. An important point to stress is that both LLMZip and FineZip use {\em teacher forcing} during compression, which allows the entire sequence to be processed in a single inference step. 
However, teacher forcing is unavailable during decompression, as the original symbols are not known in advance. Consequently, those autoregressive decoders must reconstruct each window sequentially, requiring $n$ inference steps for a window of $n$ tokens. As a result, the throughput advantage obtained from teacher forcing during compression is largely lost during decompression, which is expected to be approximately two orders of magnitude slower. In contrast, our DLM-based approach maintains nearly identical compression and decompression throughput, providing up to six orders of magnitude higher throughput than LLM-based methods during decoding, while preserving a strong compression ratio.

Having established that DLM-based compression provides a new and appealing 
compression--throughput trade-off, we now analyze the impact of the different configurations. As shown in the supplementary material (Figures 4 and 5), all compressors adopted in the last step of the DLM-based approach achieve comparable throughput --- DLM inference remains the main bottleneck --- while \texttt{lzma-9} provides the best compression ratio, with windows of $w=256-512$ tokens offering the best overall trade-offs.


\begin{figure}[!b]
    \centering
    \includegraphics[width=\linewidth]{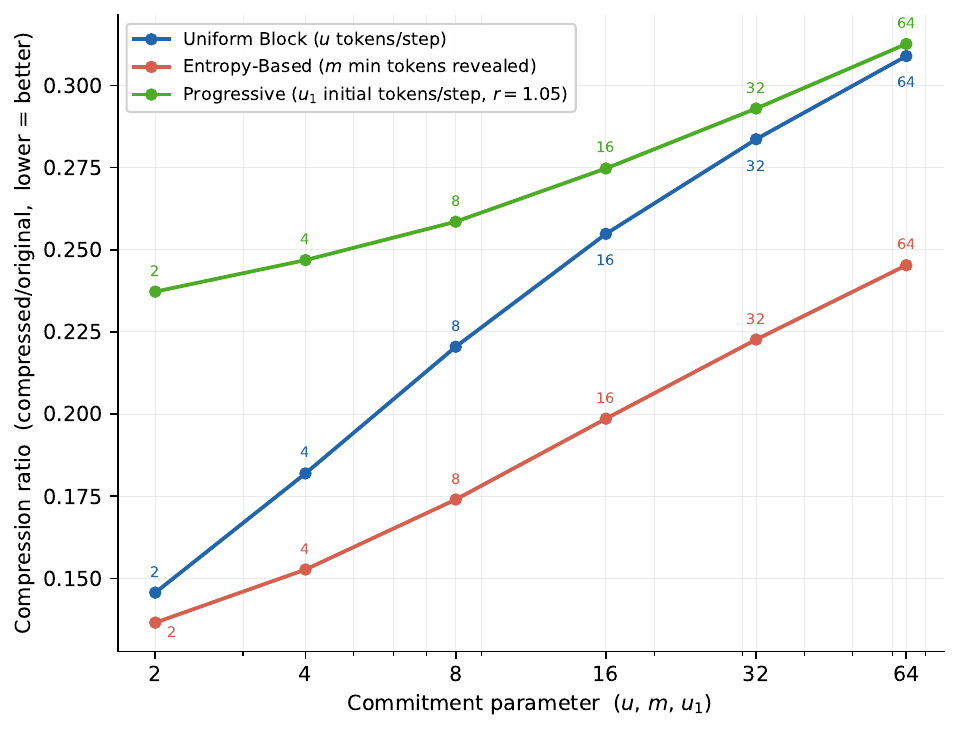}
    \caption{Compression ratio of commitment schedules as the minimum number of tokens encoded per step varies, using Symbol Ranking with $c$=2 initial context tokens.}
    \label{fig:commitment-cr}
\end{figure}

Let us now concentrate on the two key algorithmic ingredients of DLM-based compressors: commitment schedule and initial context. Regarding the commitment schedule, the results show that the entropy-based schedule achieves the best compression ratio (Figure~\ref{fig:commitment-cr}), while uniform-block provides the best trade-off between compression and throughput (Figure~7 of the supplementary material). More precisely, Figure~\ref{fig:commitment-cr} reports the compression ratio of the DLM compressor using the Symbol Ranking pipeline with $c=2$ tokens left unmasked as initial context. Each curve corresponds to a different commitment schedule, while the points along each curve represent different numbers of tokens encoded per step. Progressive is consistently the worst-performing strategy, while the entropy-based schedule achieves the best compression for every number of encoded tokens per step. Uniform-block schedule performs comparably to the entropy-based one when encoding only one or two tokens per step, but increasingly approaches the progressive schedule as more tokens are encoded per step. Figure~6 in the supplementary material corroborates these results, showing experimentally that the entropy-based schedule provides better compression at a small throughput cost compared to uniform-block, while progressive consistently achieves worse compression ratios.


Overall, these results highlight two important points: (i) relaxing the autoregressive constraint on the number of tokens encoded per step can improve compression quality; and (ii) when increasing the number of tokens encoded per step to overcome the throughput bottleneck, only entropy-based scheduling preserves a favorable compression ratio. This flexibility is unique to diffusion-based generation and unavailable to prior autoregressive compression approaches.

Finally, we investigate the effect of providing an initial context of $c$ unmasked tokens to avoid context-free first-step predictions. For space reasons, Figures~8-11 are reported in the supplementary material. Specifically, Figure~8 shows that a proper initial context improves compression across all configurations except the autoregressive one, and that its importance grows with the number of tokens encoded per step. 
Hence the initial context becomes increasingly important when targeting higher throughput. The choice of which tokens to expose is also crucial. Figures~9--11 report the average results of the adopted initial-context strategies across pipelines, commitment schedules, and numbers of tokens generated per step. The contiguous strategy, which is the only one naturally applicable to LLMs, performs well with the uniform-block schedule at low numbers of tokens generated per step, but degrades substantially at higher token counts and even more under entropy-based scheduling. At higher throughputs, it eventually performs worse than random selection! Conversely, attention-based context consistently provides the best results across all configurations.


\section{Conclusions and Future Works}

Our paper introduces the use of Diffusion Language Models (DLMs) for lossless text compression as an alternative inference paradigm to autoregressive LLM-based approaches. We show that replacing autoregressive LLMs with DLMs within the same compression framework can mitigate the throughput bottleneck imposed by their one-symbol-per-step limitation. Doing so, however, introduces novel algorithmic challenges, which we address through the design of novel commitment schedules and initial context strategies suitable for the compression setting at hand. Experiments on \texttt{enwik8}, a well-established textual benchmark, demonstrate the effectiveness of our approach against both LLM-based and general-purpose compressors.

The main message of our study is that higher throughput in neural text compression requires encoding multiple tokens per inference step and moving beyond the autoregressive structure of LLMs. In this respect, the architectural advantages of DLMs, identified and studied in this paper for the first time, proved very effective. Our experimental results are particularly promising given that DLMs are still a relatively young paradigm, with recent advances already pointing toward increasingly capable and efficient models~\cite{anagnostidis2025flexidit,kim2025autoregressive,diffusiongemma2026,vonrutte2025scaling}, thus suggesting substantial room for further improvements in diffusion-based neural compression. This potential is further amplified by the fact that 
we deployed an 8B-parameter DLM: prior work~\cite{heurtel2024compression,nardone2026llmbased} on LLMs has shown that much smaller distilled models can substantially improve the throughput with limited compression loss: a successful approach that could be investigated on DLMs too.

\bibliography{aaai2027}


\includepdf[pages=-]{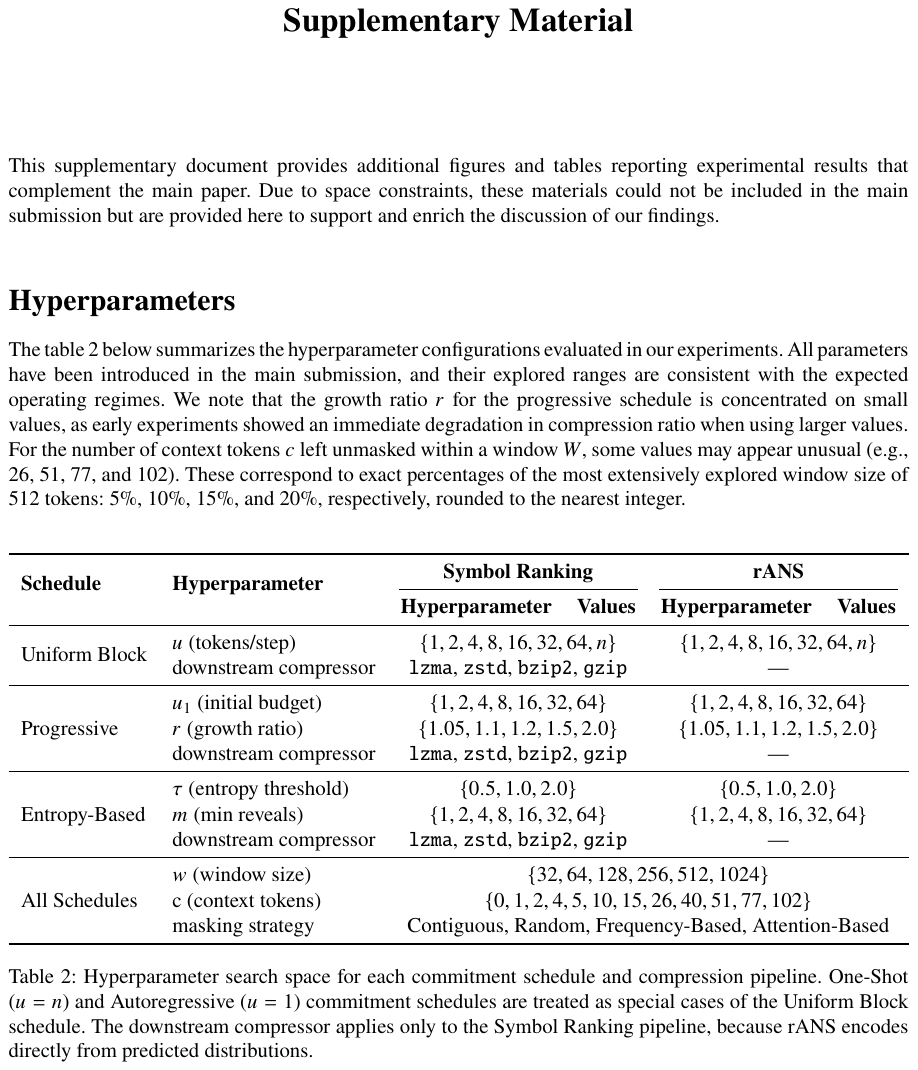}

\end{document}